\documentclass[preprint,3p,times]{elsarticle}
\biboptions{numbers,sort&compress}

\usepackage{ragged2e} 
\usepackage{booktabs,makecell, multirow, tabularx}
\usepackage{amsmath,amsfonts}
\usepackage{algorithmic}
\usepackage{algorithm}
\usepackage{array}
\usepackage[caption=false,font=footnotesize,labelfont=rm,textfont=rm]{subfig}
\usepackage{textcomp}
\usepackage{stfloats}
\usepackage{url}
\usepackage{verbatim}
\usepackage{graphicx}
\usepackage{hyperref}
\usepackage{amsthm,amsmath,amssymb,bm}
\usepackage{marvosym}
\usepackage{ifsym}
\usepackage{mathrsfs}
\usepackage{enumitem}

\makeatletter
\def\ps@pprintTitle{%
  \let\@oddhead\@empty
  \let\@evenhead\@empty
  \let\@oddfoot\@empty
  \let\@evenfoot\@oddfoot
}
\makeatother

\begin{document}

\begin{frontmatter}
\title{A Policy Gradient-Based Sequence-to-Sequence Method for Time Series Prediction}

\author[l1]{Qi Sima}
\author[l1]{Xinze Zhang}
\author[l1]{Yukun Bao\corref{cor1}}
\author[l1]{Siyue Yang}
\author[l1]{Liang Shen}

\cortext[cor1]{Corresponding author: Yukun Bao. E-mail address: yukunbao@hust.edu.cn (Y. Bao).
}

\affiliation[l1]{organization={School of Management, Huazhong University of Science and Technology},
            city={Wuhan},
            postcode={430074},
            state={Hubei},
            country={China}}
  
\begin{abstract}
Sequence-to-sequence architectures built upon recurrent neural networks have become a standard choice for multi-step-ahead time series prediction. In these models, the decoder produces future values conditioned on contextual inputs—typically either actual historical observations (ground truth) or previously generated predictions. During training, feeding ground-truth values helps stabilize learning but creates a mismatch between training and inference conditions, known as exposure bias, since such true values are inaccessible during real-world deployment. On the other hand, using the model's own outputs as inputs at test time often causes errors to compound rapidly across prediction steps. To mitigate these limitations, we introduce a new training paradigm grounded in reinforcement learning: a policy gradient-based method to learn an adaptive input selection strategy for sequence-to-sequence prediction models. Auxiliary models first synthesize plausible input candidates for the decoder, and a trainable policy network—optimized via policy gradients—dynamically chooses the most beneficial inputs to maximize long-term prediction performance. Empirical evaluations on diverse time series datasets confirm that our approach enhances both accuracy and stability in multi-step forecasting compared to conventional methods.
\end{abstract}
\end{frontmatter}
\section{Introduction}
Time series modeling and prediction are essential for automating and optimizing operational processes \citep{1995Multi}, and play a vital role in numerous real-world applications, such as advance planning of medical resources with the prediction of influenza-like illness \citep{yangComprehensiveLearningParticle2021a} and early warning of floods based on rainfall-runoff forecasts \citep{2020Exploring}. 
Multi-step-ahead time series prediction aims to estimate data values of two or more steps ahead in the future subject to a series of historical observations, which can provide more information for decision-makers than one-step-ahead prediction but is also more challenging \citep{taiebBiasVarianceAnalysis2016,2014Multi}. 

Over the last few decades, a number of statistical and machine learning methods have been proposed for multi-step-ahead time series prediction. Among these models, recurrent neural networks (RNNs) have emerged as one of the most successful techniques\citep{hewamalageRecurrentNeuralNetworks2021,yangFullScaleInformationDiffusion2023,ilhanMarkovianRNNAdaptive2023}. As a prominent architecture consisting of two independent RNNs known as encoder and decoder, sequence-to-sequence (S2S) models were originally developed for neural machine translation \citep{sutskeverSequenceSequenceLearning2014}, and have demonstrated remarkable performance in various applications \citep{zhengStochasticRecurrentEncoder2023,2020Exploring,xiangRainfallRunoffModelLSTMBased2020,salinasDeepARProbabilisticForecasting2020}. Typically, the encoder encodes the input sequences of historical observations into latent representations, while the decoder decodes and generates the output sequences of multi-step-ahead predictions. A classic approach to building S2S models is to decode out a one-step-ahead prediction, and then iteratively feed this prediction back as input to predict the next ones \citep{sangiorgioForecastingNoisyChaotic2021_FR,2020Exploring}. This autoregressive mechanism suffers from error accumulation over time as the initial errors are magnified by the feedback loop. Teacher Forcing \citep{wangImprovedWavenetNetwork2023_TF,dengDatadrivenProxyModel2021_TF}, the most prevalent training approach so far, trains the model to decode out predictions conditioned on the true values from previous time steps, rather than its own predictions. This prevents the propagation of errors and makes the output sequence closer to the target one. However, at testing stage, the decoder has to produce predictions contingent on its own predictions since true values are not available. This discrepancy between training and testing stages, known as exposure bias\citep{bengioScheduledSamplingSequence2015a}, can lead to poor performance due to an inability to copy with the buildup of errors inevitably encountered at testing stage. Consequently, developing efficient training approaches for S2S models has become a pivotal research topic and has garnered increasing attention in recent years \citep{exposurebias,goyalProfessorForcingNew2016,gasparinDeepLearningTime2019}.

The aforementioned problems would be solved if true values were available during both training and testing stages, but this is impossible. Alternatively, employing the predictions with high accuracy but from other well-trained models, hereafter referred to as auxiliary models, as inputs can effectively mitigate the dilemma, which makes it possible that the inputs could be obtained in the same way in-between training and testing stages. Furthermore, when the predictions are more accurate than that of the decoder itself, it has great potential to alleviate the further error propagation in contrast to autoregressive mechanism. In this context, selecting an optimal model from the model pool that includes auxiliary models and decoder itself, to provide inputs is therefore critical but challenging. The challenges arise as follows. Firstly, the performances of different models may differ according to the problem types, dataset characteristics, and time steps \citep{chenDynamicEnsembleWind2021a,fuReinforcementLearningBased2022,wolpertLackPrioriDistinctions1996}. Secondly, different models may have varying generalization abilities and perform differently at training and testing stages \citep{chanNeuralNetworkBasedModelsShortTerm2012}. Thirdly, the decoder is incorporated into the pool because its own one-step-ahead predictions may become more suitable as the inputs with increasing prediction accuracy. This implies that the model with the highest prediction accuracy in the pool is not static, even if only the training stage is considered. Given the complexity and variability of the foregoing factors, it is impracticable to depend on manual experience or conventional optimization methods to devise a rigid model selection scheme.

Considering that the optimal-performing model should be discerned to deliver inputs for decoder at each time step along with the decoding process, the model selection procedure can be viewed as a sequential decision problem in a dynamic and uncertain system \citep{MathematicalStatistics1967}. Driven by the rapid advancements in artificial intelligence, reinforcement learning has been proved to be a viable solution and applied successfully to many sequential decision problems \citep{1998PolciyBased,DeepReinforcementLearningTNNLS,keneshlooDeepReinforcementLearning2020,dasSolvingSemiMarkovDecision1999}. Reinforcement learning algorithms aim to build an agent that learns the optimal policy for achieving a specific goal dynamically and updates its actions adaptively through trial and error, which indicates the inherent advantage of enhancing the decoder's prediction performance by choosing appropriate inputs for the decoder in a wise way. 

In this study, a novel policy gradient-based sequence-to-sequence (PG-S2S) prediction model is proposed, which introduces an auxiliary model pool including several well-trained models to produce inputs (predictions) and then applies reinforcement learning to select the inputs. During both training and testing stages, the selected input of the optimal model from the pool is fed into the decoder to guide its decoding for the next time step. Both simulated and real-world datasets were used for the experimental analysis and to verify the validity of the proposed approach. The primary contributions of this work are summarized below:
\begin{itemize}
\item{ Introduce a novel sequence-to-sequence framework for multi-step-ahead time series prediction, integrating an auxiliary model pool and an RL-based decoding mechanism. This design explicitly addresses exposure bias and alleviates error accumulation—two key limitations of prevailing autoregressive training strategies.}
\item {Cast the choice of decoder inputs as a sequential decision-making process and solve it using a policy gradient-based reinforcement learning approach, enabling the decoder to adaptively select contextually optimal inputs at each prediction step.}
\item {Conduct extensive experiments on five real-world and well-accepted simulated time-series datasets with different time granularities. The experimental results show that our approach excels the state-of-the-art counterparts and has generalized potential for different RNN structures.}
\end{itemize}

The rest of this paper is organized as follows. Section II introduces the basic concepts related. Section III details the proposed approach. Then, the experimental setting on data, accuracy measure, and experimental procedure are described in Section IV and experimental results are discussed in Section V. Finally, section VI concludes and gives several directions of future work.
\section{Background Study}
\subsection{Multi-Step-Ahead Time Series Prediction}
Multi-step-ahead time series prediction can be described as an estimation of \(y_{t+1:t+H}\) based on the historical observations, which can be formulated as follows.
\begin{equation}
        \left\{y_{t+1},...,y_{t+H}\right\} = \mathcal{F}(Y_{t-L+1},...,Y_{t})+\varepsilon
\end{equation}
where \(H\) is an integer greater than one, known as forecasting horizon. \((Y_{t-L+1},...,Y_{t})\) is the model input vector consisting of historical observations, where \(L\) is the number of time lags. If \(Y_{t-L+1:t} = y_{t-L+1:t}\), \emph{i.e.}, the input only contains historical observations of the target variate \(y\), it is referred to as a univariate prediction problem. While \(Y_{t-L+1:t}=\left[y_{t-L+1:t},x_{t-L+1:t}^1,...,x_{t-L+1:t}^n\right]^{T}\), \emph{i.e.}, the input contains historical observations of external variates \(\left\{x^1,...,x^n\right\}\) in addition to the target variate, it is referred to as a multivariate prediction task. \(\mathcal{F}(\cdot)\) is the function approximated by the model developed for the problem with the input. \(\varepsilon\) denotes the error associated with the function \(\mathcal{F}\).

Many scholars have been studying the problem and proposed a variety of models for multi-step-ahead prediction. Machine-learning models, such as multiple-output support vector regression (MSVR) and neural networks (NNs), can model more complex dependencies than traditional statistical models like Naive \citep{Naive} and autoregressive integrated moving average (ARIMA) \citep{1970ARIMA}, and thus have become the mainstream modelling techniques. MSVR is an extension of the standard SVR that can generate multiple outputs at once by mapping input data into a high-dimensional space and performs regression by finding an optimal hyperplane \citep{Tuia2011MSVR}. It has low computational complexity, but struggles with large-scale data, which can be handled by NNs. Typical NNs include multi-Layer perceptron (MLP) and recurrent neural networks. MLP \citep{yangComprehensiveLearningParticle2021a,hewamalageRecurrentNeuralNetworks2021} is the most basic neural network model that consists of an input layer, one or more hidden layers and an output layer. It has good nonlinear approximation capabilities but cannot capture the temporal correlation among input data. RNNs, on other hand, have shown superior capability on capturing temporal dependencies. They take the sequence data as inputs and consist of a number of units sharing weight parameters and recursively connecting in the evolutionary direction of the sequence in a chain-like manner. The most popular recurrent units are ERNN \citep{1990vanillaRNN}, LSTM \citep{1997Lstm}, and GRU \citep{2014gru}. In this study, we take LSTM with S2S structure to illustrate the mechanism of the proposed PG-S2S and the novel framework to achieve better multi-step-ahead predictions. Note that we use the term ``S2S" alternatively more often than LSTM in places where the structure of inputs should be highlighted.
\subsection{Reinforcement Learning}
Reinforcement learning (RL) is inspired by behaviorist psychology and aims to mimic the human learning ability to select actions that maximize long-term profit in the interactions with the environment. The RL has been around for decades and applied to various sequential decision-making problems. Specifically, RL defines any decision maker (learner) as an \emph{agent} and anything outside the agent in the problems as an \emph{environment} \citep{1998PolciyBased}.  Before using RL algorithms, it is usually necessary to model the problem as a Markov decision process (MDP), which comprises a state space \(\mathcal{S}\), an action space \(\mathcal{A}\), a state transition function \(\mathcal{P}\) satisfying the Markov property, and a reward calculation function \(\mathcal{R}\). The agent learns from feedback by interacting with the environment, where the interaction can be described as follows. At each step \(k\) of the sequential decision problems, the agent perceives the current state \(s_{k}\) of the environment and selects an action \(a_{k}\) according to a mapping from states to actions called the \emph{policy}. The environment then alters its state based on transition function, and provides a feedback reward to the agent. 

Within the framework of RL, the goal of sequential decision-making is shifted from finding an optimal set over all possible sequences of actions to focusing only on the optimal actions to take in each state, \emph{i.e.}, learning the optimal policy. Therefore, it can lead to more efficient and effective decision-making in complex, dynamic input selection problems of the decoder where traditional methods may be limited. Among the categories of RL algorithms, variants of policy gradient (PG) algorithms have consistently produced state-of-the-art results \citep{PolciyG2019}, which model the policy directly with a parameterized machine learning model. In general, the policy \(\pi_{\theta}\) is a probability distribution over actions conditioned on state. For the case of stochastic policy where the action space is discrete such as the input selection problems in this study, the policy can be defined as follows. 
\begin{equation}
        \pi_{\theta}(a|s) = P[a|s;\theta].
\end{equation}
\indent In this way, the optimal policy is obtained by optimizing the parameters \(\theta\) using gradient ascent with a set of training samples, which is theoretically supported by the \emph{stochastic policy gradient theorem} \citep{1998PolciyBased,PGTheorem1999}. Among PG algorithms derived by the policy gradient theorem, REINFORCE is almost the most representative one \citep{1992MCPG,2014DPG,mainPG2016}, which is followed to give the calculation formula of the gradient in this study, as shown in section III. 
\begin{figure*}[!t]
    \begin{minipage}[t]{0.66\textwidth}
        \centering
        \includegraphics[width=\textwidth,height=6.8cm]{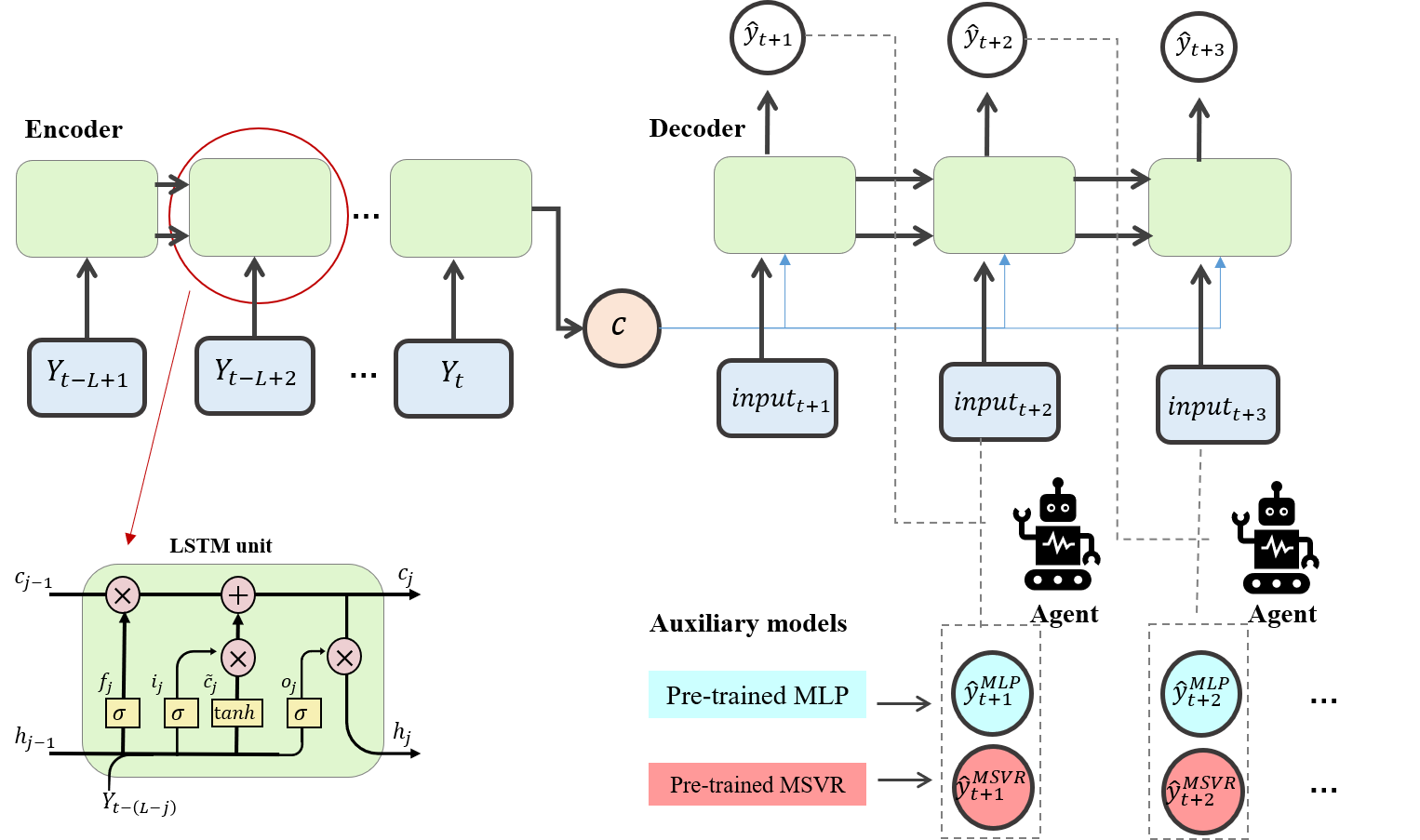}
        \caption{Basic framework of S2S model with policy gradient.}
        \label{figure:RL_Decoder}
    \end{minipage}%
    \begin{minipage}[t]{0.32\textwidth}
        \centering
        \includegraphics[width=\textwidth,height=6.5cm]{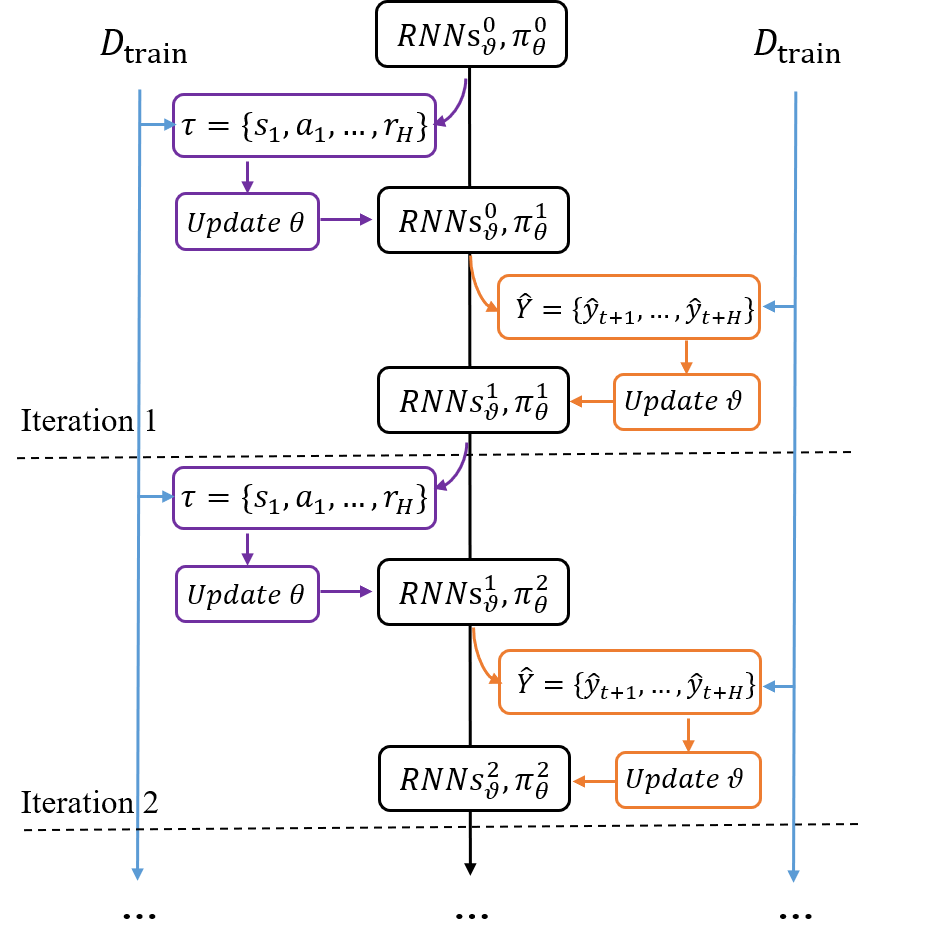}
        \caption{Illustration of asynchronous training.}
        \label{figure:Training}
    \end{minipage}
\end{figure*}
\section{Methodology}
In the application of the S2S model for multi-step-ahead time series prediction, the input sequence is composed of 
historical observations denoted by \(Y_{t-L+1:t}\) in section II, while the output sequence represents estimations of future values denoted by \(y_{t+1:t+H}\). This section firstly presents the MDP formulation for input selection of the decoder. Subsequently, the proposed PG-S2S model is described, and its training and construction processes are explained.
\subsection{MDP Setting for Input Selection of the Decoder} The problem involves a sequence of model selection decisions along the decoding process, with the goal of enhancing the multi-step-ahead prediction performance of the decoder, where the predictions of the selected model are fed as inputs to the decoder. Each decision affects the subsequent choices and outcomes, thus forming a sequential decision problem where the decoder serves as the environment. As illustrated in Fig. \ref{figure:RL_Decoder}, at each time step \(k\in\left\{1,2,...,H\right\} \), the agent observes the current decoding situation and selects a model whose prediction is fed into the decoder. The decoder then transits to the next state and gives a reward to the agent. The specific mathematical model, MDP \(\left\langle \mathcal{S},\mathcal{A},\mathcal{P},\mathcal{R}\right\rangle\) is summarized as follows.

\textbf{State-space} \(\bm{\mathcal{S}}\). Each state \(s_{k}\) is a vector that characterizes the current situation of the environment at time step \(k\). Here \(\mathcal{S}\) is a continuous state space, which is naturally defined as the hidden state of the decoder since it stores the decoding status that retains all relevant information for the agent to select actions. 

\textbf{Action-space} \(\bm{\mathcal{A}}\). Each action \(a_{k}\) describes the behavioral characteristics of agent at time step \(k\), which is an element in \(\mathcal{A}\). Here the action space is a set containing the models selected to
produce the predictions (inputs) for the decoder, \emph{i.e.}, the decoder itself and the auxiliary models.

\textbf{Transition-function} \(\bm{\mathcal{P}}\). Since state \(s_{k}\) is defined as hidden-state of the decoder, whose transitions follow the rules of RNN, \emph{i.e.}, \(s_{k+1}=\text{\emph{RNNs}}_{\vartheta}(s_{k},a_{k})\). \(\text{\emph{RNNs}}_{\vartheta}\) includes encoder RNN and decoder RNN with parameters \(\vartheta\), consisting of weight matrices and bias vectors related to the forget, input, and output gates and output layer detailed in Section III.B.

\textbf{Reward-function} \(\bm{\mathcal{R}}\) \citep{fuReinforcementLearningBased2022}. Reward \(r_k\) is a numerical value representing the instant reward to the agent affiliated with executing action \(a_{k}\) in state \(s_{k}\), which is calculated by \(\bm{\mathcal{R}}\). As the agent learns by trial and error, the design of the reward function dictates the optimal policy. In this study, it is desired that the decoder generates accurate predictions for the next time step after being fed with the predictions from the model selected by the agent. Therefore, \(r_{k}\) is defined as follows.
\begin{equation}
        r_k = \alpha\cdot \text{\emph{Rank\_}}r_{k} + (1-\alpha)\cdot \text{\emph{Accuracy\_}}r_{k} 
\end{equation}
where \(\alpha\) is a hyperparameter balancing the two type rewards. \(\text{\emph{Rank\_}}r_{k}\) is designed to motivate the agent select the model with the highest prediction accuracy at step \(k\), while \(\text{\emph{Accuracy\_}}r_{k}\) evaluates the performance of the decoder at step \(k+1\) after using the prediction of the model chosen by the agent as the input. Hence, they are defined as: 
\begin{equation}
        \text{\emph{Rank\_}}r_{k}= 1-\frac{rank\left(a_k\right)}{N^{a}}
\end{equation} 
where \(N^{a}\) is the number of models available in the model pool. The models are rank ordered from 1 (the best) to \(N^{a}\) (the worst) by prediction accuracy and \(rank\left(a_k\right)\) indicates the ordinal number of the model selected by the agent at step \(k\).
\begin{equation}
        \text{\emph{Accuracy\_}}r_{k}=\frac{\beta}{\beta + \left\lvert y_{t+k+1}-\hat{y}_{t+k+1} \right\rvert }
\end{equation}
where \(\beta\) is a hyperparameter converting penalties to reward-based metrics, \(\hat{y}_{t+k+1}\) is the prediction of the decoder at step \(k+1\) after being fed into predictions selected. Note that \(\text{\emph{Accuracy\_}}r_{H}=0\).
\subsection{The PG-based S2S Prediction Model}
Compared to the standard S2S with only encoder and decoder, the proposed PG-S2S has the agent and auxiliary models besides. For the sake of simplicity but without loss of generality, LSTM unit (as shown in Fig. \ref{figure:RL_Decoder}) is used to constitute the RNNs of both encoder and decoder because it can capture long-term dependencies in the sequence, alleviate gradient vanishing issues and have good performance in many forecasting tasks \citep{hewamalageRecurrentNeuralNetworks2021,2020Exploring,xiangRainfallRunoffModelLSTMBased2020}.

\textbf{Encoder} is to map the sequence of the historical observations \(Y_{t-L+1:t}\) to a fixed-sized (context) vector, and can be expressed as:
\begin{align}
        \label{equation:encoder_1}
        &f_{j} = \sigma \left( U_{f}^{e}\left[h_{j-1},input_{j}\right]+b_{f}^{e}\right)\\
        &i_{j} = \sigma \left( U_{i}^{e}\left[h_{j-1},input_{j}\right]+b_{i}^{e}\right)\\
        &\tilde{c}_{j} = tanh\left(U_{c}^{e}\left[h_{j-1},input_{j}\right]+b_{c}^{e}\right)\\
        &c_{j} = i_t\odot \tilde{c}_{j} + f_{j} \odot c_{j-1} \\
        &o_{j} = \sigma \left( U_{o}^{e}\left[h_{j},input_{j}\right]+b_{o}^{e}\right)\\
        &h_{j} = o_{j}\odot tanh\left(c_{j}\right)
        \label{equation:encoder_2}
\end{align}
where \(h_{j},c_{j}\in \mathbb{R}^{N^{e}}\) are the hidden state and the cell state of encoder at step \(j\in\left\{1,2,...,L\right\}\), corresponding to the short-term memory and the long-term memory respectively. \(N^{e}\) is the dimension of LSTM cells in encoder. While \(\tilde{c}_{j}\in \mathbb{R}^{N^{e}}\) is the candidate cell state used for temporary storage of information. \(input_{j}\in \mathbb{R}^m\) is the encoder input at step \(j\) and \(m\) is the number of variates, which equals to \(Y_{t-(L-j)}\). Moreover, \(U_{f}^{e},U_{i}^{e},U_{o}^{e}\in \mathbb{R}^{N^{e}\times\left(N^{e}+m\right)}\) and \(b_{f}^{e},b_{i}^{e},b_{o}^{e}\in \mathbb{R}^{N^{e}}\) denote the weight matrices and bias vectors of the forget gate, input gate, and output gate, respectively. \(f_{j},i_{j},o_{j}\in \mathbb{R}^{N^{e}}\) represent the corresponding gate vectors, marking the information to be discarded, stored and outputted. \(\sigma(\cdot)\) and \(tanh(\cdot)\) are activation functions. 

Then the hidden states up to step \(L\) are built according to equation \eqref{equation:encoder_1}-\eqref{equation:encoder_2}, where the last one \(h_{L}\), summarizing the whole sequence of historical observations, is usually used as the context vector \textbf{\emph{c}} passed to the decoder \citep{xiangRainfallRunoffModelLSTMBased2020,Wen2017_quantile}.

\textbf{Decoder} is to generate the output sequence, \emph{i.e.}, the estimation of \(y_{t+1:t+H}\), combined with the context vector and the inputs from the models selected by the agent, which can be defined as:
\begin{align}
        &f_k = \sigma \left( U_{f}^{d}\left[s_{k-1},input_{t+k},\textbf{\emph{c}}\right]+b_{f}^{d}\right)\\
        &i_k = \sigma \left( U_{i}^{d}\left[s_{k-1},input_{t+k},\textbf{\emph{c}}\right]+b_{i}^{d}\right)\\
        &\tilde{c}_{k} = tanh\left(U_{c}^{d}\left[s_{k-1},input_{t+k},\textbf{\emph{c}}\right]+b_{c}^{d}\right)\\
        &c_k = i_{k}\odot \tilde{c}_k + f_t \odot c_{k-1}\\
        &o_k = \sigma \left( U_{o}^{d}\left[s_{k-1},input_{t+k},\textbf{\emph{c}}\right]+b_{o}^{d}\right)\\
        &s_k = o_k\odot tanh\left(c_k\right) \\
        &\hat{y}_{t+k}  = V^{d}h_{k}+b^{d}   
\end{align}
where \(s_{k},c_{k},\tilde{c}_{k}\in \mathbb{R}^{N^{d}}\) are the hidden state, the cell state, and candidate cell state of decoder at step \(k\in\left\{1,2,...,H\right\}\). \(N^{d}\) is the cell dimension. Likewise, \(U_{f}^{d},U_{i}^{d},U_{o}^{d}\in \mathbb{R}^{N^{d}\times\left(N^{d}+1+N^{e}\right)}\), \(b_{f}^{d},b_{i}^{d},b_{o}^{d}\in \mathbb{R}^{N^{d}}\), \(f_{j},i_{j},o_{j}\in \mathbb{R}^{N^{d}}\) denote the matrices and vectors related to the forget, input, and output gates of decoder, respectively. \(V^{d} \in \mathbb{R}^{1\times N^d},b^{d}\in \mathbb{R}^1\) denote the weight matrix and bias vector of output layer. Note that \(input_{t+k}\in \mathbb{R}^{1}\) is the decoder input at step \(k\), which is the prediction of \(y_{t+k-1}\) produced by model selected by agent from model pool when \(k>1\), and is the ground-truth value \(y_{t}\) when \(k=1\), \emph{i.e.},
\begin{equation*}
        {input_{t+k}} = 
        \begin{cases}
                y_{t},  & {k=1} \\
                {\hat{y}_{t+k-1}^{a_{k}},} & {k>1.}
        \end{cases}
\end{equation*}

Based on the PG algorithm, the policy \(\pi_{\theta}\) of the agent is modelled directly by a single hidden-layer MLP with softmax activation function as defined below \citep{DBLP_MLP}. The parameters \(\theta\) encompass the weight matrices and bias vectors in MLP.
\begin{equation}
        \pi_{\theta}(s) \triangleq softmax\left(W_2\left(\sigma\left(W_1s+b_1\right)\right)+b_2\right) 
\end{equation}
where \(W_1\in\mathbb{R}^{N^{p}\times N^{d}},W_2\in\mathbb{R}^{ N^{a}\times N^{p}}, b_1\in\mathbb{R}^{N^{p}},b_2\in\mathbb{R}^{N^{a}}\) denote the weight matrices and bias vectors of hidden layer and output layer; \(N^{p}\) is the number of nodes in the hidden layer. \(N^{a} = 3\), which is also the number of actions in \(\mathcal{A} =\left\{\text{\emph{MSVR}},\text{\emph{MLP}},\text{\emph{Decoder}}\right\}\). Note that we consider two auxiliary models in this study, \emph{i.e.}, well-trained MSVR and MLP. 

Let \(\hat{o}=[P(a_1),P(a_2),P(a_3)]^{T}\) denotes the output of the policy network, where each element represents the selection probability of the corresponding action. Then the action (model) selected by the agent in the state \(s_k\) is obtained by: 
\begin{equation}
        \label{equation:agent}
        a_{k} = \underset{a \in \mathcal{A}}{\arg\max } \, \pi_\theta\left(a \mid s_{k}\right). \\
\end{equation}
\subsection{Training and Prediction}
\textbf{Agent}. Let \(\tau = \left\{s_1,a_1,r_1,...,s_H,a_H,r_H\right\}\) be a trajectory denoting the interactions between agent and environment in one complete decoding process. The agent's goal is to obtain a policy which maximizes the expected total reward along \(\tau\), defined as:
\begin{equation}
        \begin{aligned}
                \max J\left(\pi_{\theta}\right) 
                & = E_{\tau \sim \pi_{\theta }}\left[R(\tau)\right]
        \end{aligned} 
\end{equation}
where \(R(\tau)=\sum_{k=1}^{H}\gamma^{k}r_k\) is the cumulative discounted reward from the start state, \(\gamma\) is a discounted factor so that \(0\leq\gamma<1\). \(P\left(\tau \right)\) denotes the probability of producing \(\tau\) given the policy \(\pi_{\theta}\) and encoder-decoder \(\text{\emph{RNNs}}_{\vartheta}\), \emph{i.e.}, \(P\left(\tau \right)=P_{\text{\emph{RNNs}}_{\vartheta }}\left(s_1\right)\prod_{k=1}^{H}\pi_\theta\left(a_k \vert s_k\right)P_{\text{\emph{RNNs}}_{\vartheta }}\left(s_{k+1}\vert s_k,a_k\right)\). The gradient of training object for agent is formulated as:
\begin{equation}
        \label{euq:gradient1}
        \begin{aligned}
        \bigtriangledown_{\theta} J\left(\pi _{\theta }\right)
        &=\bigtriangledown_{\theta}\displaystyle\sum_{\tau }P\left(\tau \right)R\left(\tau\right)\\
        &= \displaystyle\sum_{\tau } P\left(\tau \right)\bigtriangledown_{\theta}logP\left(\tau \right)R\left(\tau\right).
        \end{aligned}
\end{equation}
\begin{algorithm}[!t]
    \caption{The PG-based S2S Modeling Algorithm.}
    \label{alg:alg1}
    \begin{algorithmic}[1]
            \STATE \textbf{Input:} the supervision samples $D_{train}$, well-trained auxiliary models, the hyper-parameters of RNNs and policy, and the learning rate \(l_1,l_2\).
            \STATE Initialize the policy network $\pi_\theta$, and the RNNs of encoder and decoder $\text{\emph{RNNs}}_{\vartheta}$ with random parameters $\theta,\vartheta$.
            \STATE \textbf{While} not sufficiently good performance or maximum number of iterations \textbf{do}
            \STATE \hspace{0.26cm} Fix parameters $\vartheta$ of RNNs.
            \STATE \hspace{0.2cm} Collecte a set of trajectories by executing the current policy $\pi_\theta$ with \(\epsilon\)-greedily strategy.
            \STATE \hspace{0.2cm} \textbf{For} each trajectory \(\tau = \left\{s_1,a_1,r_1,...r_H\right\}\) \textbf{do}
            \STATE \hspace{0.6cm} \textbf{For} \(k \in \left\{1,..,H\right\} \) \textbf{do}
            \STATE \hspace{1.0cm} Calculate $G_k=\sum_{k'= k}^{H}\gamma^{k'- k}r_{k'}$.
            \STATE \hspace{1.0cm} Update policy: $\theta \leftarrow \theta + l_1\gamma^{k}G_k\bigtriangledown_{\theta}log\pi_{\theta }(a_k\mid s_k)$.
            \vspace{-4mm}
            \STATE \hspace{0.6cm} \textbf{Endfor}
            \STATE \hspace{0.2cm} \textbf{Endfor}
            \STATE \hspace{0.2cm} Fix parameters $\theta$ of policy.
            \STATE \hspace{0.2cm} Calculate a set of prediction sequences by executing the current encoder and decoder $\text{\emph{RNNs}}_{\vartheta}$ as equations \eqref{equation:encoder_1}-\eqref{equation:agent}.
            \STATE \hspace{0.2cm} \textbf{For} each multi-step-ahead prediction sequence $\hat{Y}=\hat{y}_{t+1:t+H}$ \textbf{do}
            \STATE \hspace{0.6cm} Calculate $\mathcal{L}(\hat{Y})=\frac{1}{H}\sum_{k = 1}^{H}(y_{t+k} - \hat{y}_{t+k})^{2}$.
            \STATE \hspace{0.6cm} Update RNNs: $\vartheta \leftarrow \vartheta - l_2\bigtriangledown_{\vartheta}\mathcal{L}(\hat{Y})$.
            \STATE \hspace{0.2cm} \textbf{Endfor}
    \end{algorithmic}
    \label{alg1}
\end{algorithm}
As asynchronous training is used, the RNNs remain unchanged when the policy is updated, which is independent of \(\theta \). Both \(\bigtriangledown_{\theta}logP_{\text{\emph{RNNs}}_{\vartheta}}(s_1)\) and \(\bigtriangledown_{\theta}logP_{\text{\emph{RNNs}}_{\vartheta}}(s_{k+1}\mid s_k,a_k)\) are equal to 0. Then equation \eqref{euq:gradient1} can be transferred as follows.
\begin{equation}
        \begin{aligned}
        \bigtriangledown_{\theta} J\left(\pi _{\theta }\right)
        &= \sum_{\tau } P\left(\tau \right)\sum_{k = 1}^{H}\bigtriangledown_{\theta}log\pi_\theta (a_k\mid s_k)R\left(\tau\right)\\
        &= E_{\tau \sim \pi_{\theta }}\sum_{k = 1}^{H}\bigtriangledown_{\theta}log\pi_\theta(a_k\mid s_k)R\left(\tau\right)\\
        &= \frac{1}{N}\sum_{i = 1}^{N}\sum_{k = 1}^{H}\bigtriangledown_{\theta}log\pi_{\theta}(a_k^i\mid s_k^i)R\left(\tau^{i}\right).
        \end{aligned}
\end{equation}

Besides, we take \(\epsilon\)-greedily exploration strategy to execute the actions to jump out of the local optimal point during the training process, where the agent executes an action randomly with a certain probability \(\epsilon\) less than one, instead of the policy-based action, \emph{i.e.},
\begin{equation*}
        {a_{k}} = \begin{cases}
                \underset{a \in A}{\arg\max } \, \pi_\theta\left(a \mid s_k\right), & 1-\epsilon \\
                random ~ a \in A, & \epsilon.
        \end{cases}
\end{equation*}

The agent does not explore and acts exclusively according to the policy like equation \eqref{equation:agent} when \(\epsilon = 0\).

\textbf{RNNs}. While the parameters of the encoder and the decoder (\emph{i.e.}, RNNs), are trained to minimize the deviation of the ground-truth sequence from prediction sequence, that is,
\begin{equation}
        \min J\left(\text{\emph{RNNs}}_{\vartheta}\right) = E_{\hat{Y} \sim \text{\emph{RNNs}}_{\vartheta }}[\mathcal{L}(\hat{Y})]
\end{equation}
where \(\mathcal{L}(\hat{Y})=\frac{1}{H}\sum_{k = 1}^{H}(y_{t+k} - \hat{y}_{t+k})^{2}\), is the Mean Square Error, 
then the gradient of training object for \(\text{\emph{RNNs}}_{\vartheta}\) is formulated as:
\begin{equation}
        \begin{aligned}
        \bigtriangledown_{\vartheta} J\left(\text{\emph{RNNs}}_{\vartheta }\right) 
        &=\frac{1}{N}\sum_{i = 1}^{N}\bigtriangledown_{\vartheta}\mathcal{L}(\hat{Y}^i)\\
        &=\frac{2}{NH}\sum_{i = 1}^{N}\sum_{k = 1}^{H}(\hat{y}_{t+k}^i-y_{t+k}^i)\bigtriangledown_{\vartheta}\hat{y}_{t+k}^i.
        \end{aligned}
\end{equation}
\section{Experiments}
\subsection{Datasets Description}
In this study, we evaluate the performance of the proposed PG-S2S on five datasets that cover both simulated and real-world time series at different time granularities, and in type of either univariate or multivariate. The simulated time series is the Mackey-Glass time series (\textbf{MG}), while the real-world time series come from four typical time series prediction scenarios: temperature prediction (\textbf{SML}), influenza-like illness rates prediction in the medicine field (\textbf{ILI}), \(\text{PM}_{2.5}\) prediction in environmental protection (\textbf{PM}), and prediction of operating indicators for electricity equipment in the energy sector (\textbf{ETT}). A brief description of each series is provided below, and the prediction tasks are summarized in Table \ref{tab:tables.data_Info.tex}.
\begin{table}[t!]
    \centering
    \footnotesize
    \caption{Information of Datasets and Prediction Tasks}
    \setlength{\tabcolsep}{1pt}
    \begin{tabular}{cccccc}
        \toprule
        \multirow{2}{*}{Dataset}  & Simulated   & \multicolumn{4}{c}{Real-world}  \\ 
        \cmidrule(l){2-2} \cmidrule(l){3-6} 
        & MG  & SML  & PM & ILI & ETT    \\ 
        \midrule
        No. of time series  & 1		& 1	    & 1 	& 1  	& 2 \\
        Frequency   & -		& minutely	& daily	    & weekly	& hourly \\
        Length of time series   & 7,000 	& 2,764  & 2,191  & 629   & 4,244 \\
        Time lags \(d\)   & 200 	& 288  & 180  & 26   & 168 \\
        Prediction horizon \(H\)   & 17,84 	& 48,96  & 30,60  & 4,12   & 24 \\
        Characteristics   & Univariate 	& Univariate  & Univariate  & Univariate   & Multivariate \\
        \bottomrule
    \end{tabular}
    \label{tab:tables.data_Info.tex}
\end{table}

\textbf{MG} \citep{2014Multi}. The Mackey-Glass time series has been commonly recognized as the benchmark in a series of studies related to time series prediction, which is usually generated by the nonlinear delay differential equation as follows.
\begin{equation}
        \frac{d x_{t}}{d t}=\frac{0.2 x_{t-17}}{1+x^{10}_{t-17}}+0.1 x_{t}
\end{equation}
where \(x_t\) is initialized as 1.2 with reference to the setting of \citep{2014Multi}, and 7000 observations are simulated.

\textbf{SML}\footnote{https://archive.ics.uci.edu/ml/datasets/sml2010}. The SML time series was collected from an indoor temperature monitoring system installed in a domotic house over a period of approximately 40 days. The data was sampled every minute and then calculated and smoothed with 15 minutes means. 

\textbf{PM}\footnote{https://archive.ics.uci.edu/ml/datasets/PM2.5+Data+of+Five+Chinese+Cities}. This daily time series contains the \(\text{PM}_{2.5}\) data in Chengdu from January 1st, 2010 to December 31st, 2015, obtained by averaging the hourly data in the publicly available dataset \citep{Shuyi2016PM}. 

\textbf{ILI} \citep{yangComprehensiveLearningParticle2021a}. The time series  of weekly influenza-like illness rates\footnote{https://github.com/XinzeZhang/TimeSeriesForecasting-torch/tree/ master/data/real/ili} was collected from Influenza Weekly Report issued on the website of Chinese National Influenza Center, containing 629 values observed in southern China from the first week of 2010 to the third week of 2022. The rate is defined by: 
\begin{equation}
        \text{\emph{ILI rate}} = \frac{N_{\text{\emph{ILI}},t}}{N_{v,t}}
\end{equation}
where \(N_{\text{\emph{ILI}},t}\) is the number of weekly ILI hospital visits and \(N_{v,t}\) is the number of total weekly hospital visits.

\textbf{ETT} \citep{WOS:000681269802090,WOS:000901616402073}. The ETT dataset consists of data collected from electricity transformers in two regions of a province in China. Each data point contains the target variable 'oil temperature (OT)' and six different types of external power load features. In this study, two time series at the hourly level from 2018-01-01, namely ETTh1 and ETTh2, are used for the experiment.
\subsection{Accuracy Measure}
Given a \(H\)-step-ahead prediction sequence \(\left\{\hat{y}_{t + 1},...,\hat{y}_{t+H}\right\}\) and the corresponding ground truth sequence \(\left\{y_{t + 1},...,y_{t+H}\right\}\), three prediction accuracy measures are considered: the Root Mean Square Error (RMSE), the Mean Absolute Percentage Error (MAPE), and the Symmetric Mean Absolute Percentage Error (SMAPE). The definitions of each are as follows \citep{2020Error}.
\begin{equation*}
        \begin{aligned}
                \text{\emph{RMSE}} 
                &= \sqrt{\frac{1}{H}\sum_{k = 1}^{H}(y_{t+k} - \hat{y}_{t+k})^{2}}. \\
                \text{\emph{MAPE}} 
                &= \frac{1}{H}\sum_{k = 1}^{H}\left\lvert \frac{y_{t+k}-\hat{y}_{t+k}}{ y_{t+k} }\right\rvert. \\
                \text{\emph{SMAPE}} 
                &= \frac{1}{H}\sum_{k = 1}^{H}\frac{\left\lvert y_{t+k}-\hat{y}_{t+k}\right\rvert }{\left\lvert  y_{t+k}+\hat{y}_{t+k}\right\rvert }. 
        \end{aligned}
\end{equation*}
\subsection{Experimental Setup}
In the experiment, we compare the performance of the proposed model with four well-established training approaches including free running, teacher forcing, scheduled sampling, and professor forcing, to validate the effectiveness of the PG-S2S. Additionally, well-trained MLP and MSVR models are also included as comparisons, as they are parts of the input model pool of the PG-S2S. A Naive model, which employs the historical observation at the last time step as its prediction values, acts as the baseline.

\textbf{Free Running (FR)}. It is a traditional training approach in which the predictions generated early in the decoding process are fed back as inputs for the generation of predictions at later steps. That is, both training and testing are based on the autoregressive mechanism to generate multi-step-ahead predictions.

\textbf{Teacher Forcing (TF)}. It is a popular training approach, where the teacher is the ground-truth sequence. During training, ground-truth values are fed as inputs for the generation of predictions at later steps. During testing, however, the autoregressive mechanism is used since there are no teachers (ground-truth sequence) available.

\textbf{Scheduled Sampling (SS)} \citep{bengioScheduledSamplingSequence2015a}. It is a curriculum learning strategy that gradually converts the training process from a fully guided scheme using ground-truth values to a less guided scheme that mostly uses generated predictions instead. This approach is an improvement over teacher forcing and is designed to reduce exposure bias. At every time step, a coin is flipped to decide whether to use the previous ground-truth values with probability \(p\) or one sampled from the decoder itself with probability \((1-p)\) during training. Generally, \(p\) is larger in the early stages of training and gradually decreases as the network becomes more fully trained, thus reducing reliance on ground-truth values. This keeps the training and testing as consistent as possible. There are many decay schedules for \(p\), including linear decay schedule, exponential decay schedule and so on. In this study, we followed the linear decay schedule in accordance with \citep{bengioScheduledSamplingSequence2015a, muralidharMultivariateLongTermState2019}.

\textbf{Professor Forcing (PF)} \citep{goyalProfessorForcingNew2016}. It is another typical improvement over teacher forcing to reduce exposure bias by applying adversarial training techniques. In this approach, there are two shared-weight S2S models that serve as generators: one decodes and generates output sequences using free running, while the other decodes and generates output sequences using teacher forcing. In addition to generators, there is a distinct neural network that functions as a discriminator whose objective is to accurately distinguish whether the distributions of the hidden states are derived from free running or teacher forcing. The S2S models are optimized with the aim of fooling the discriminator while minimizing prediction error.

\begin{table*}[!t]
    \footnotesize
    \caption{Results of All Methods on the Five Datasets}
    \centering
    \footnotesize
    \begin{tabular}{ccccccccccc}
    \toprule
    \multirow{2}{*}{Metric} & \multirow{2}{*}{Dataset} & \multirow{2}{*}{H} &  \multirow{2}{*}{Naive} & \multicolumn{2}{c}{Auxiliary models} & \multicolumn{5}{c}{S2S models with different training approaches} \\  \cmidrule(l){5-6} \cmidrule(l){7-11} 
    &  & & &MSVR &MLP &\makecell{FR }&\makecell{TF} &\makecell{SS} & \makecell{PF} & \makecell{PG-S2S} \\
    \midrule

    \multirow{10}{*}{RMSE} 
    & \multirow{2}{*}{MG} & 17  
    &2.66E-01	&8.79E-03	&4.80E-03	&5.00E-03	&1.56E-02	&6.10E-03	&\underline{1.38E-03}	&\underline{\textbf{1.25E-03}}  \\
    \cmidrule(l){3-11}
    &   & 84 
    &3.30E-01	&1.52E-02	&\underline{5.98E-03}	&8.78E-02	&1.99E-01	&1.05E-01	&1.14E-01	&\underline{\textbf{2.55E-03}}    \\
    \cmidrule(l){2-11}

    & \multirow{2}{*}{SML} & 48 
    &2.67E+00	&1.55E+00	&\underline{1.40E+00}	&1.73E+00	&2.22E+00	&1.79E+00	&2.10E+00	&\underline{\textbf{1.04E+00}}\\
    \cmidrule(l){3-11}
    &   & 96
    &2.73E+00	&1.59E+00	&\underline{1.46E+00}	&2.26E+00	&2.29E+00	&2.26E+00	&2.68E+00	&\underline{\textbf{1.10E+00}}\\
    \cmidrule(l){2-11}

    & \multirow{2}{*}{PM} &30 
    &4.55E+01	&3.93E+01	&3.92E+01	&3.69E+01	&\underline{3.68E+01}	&3.72E+01	&3.73E+01 &\underline{\textbf{3.59E+01}}\\
    \cmidrule(l){3-11}
    &   & 60 
    &5.14E+01	&4.05E+01	&\underline{3.73E+01}	&3.95E+01	&4.00E+01	&3.96E+01	&3.96E+01&\underline{\textbf{3.61E+01}}\\
    \cmidrule(l){2-11}

    & \multirow{2}{*}{ILI} & 4 
    &7.32E-01	&7.30E-01	&8.66E-01	&\underline{6.71E-01}	&6.73E-01	&6.72E-01	&6.75E-01 &\underline{\textbf{6.68E-01 }}\\
    \cmidrule(l){3-11}
    &   & 12 
    &1.20E+00	&\underline{9.75E-01}	&1.03E+00	&1.02E+00	&1.02E+00	&1.03E+00	&1.03E+00 &\underline{\textbf{9.71E-01 }}\\
    \cmidrule(l){2-11}
    & ETTh1  & 24
    &1.63E+00	&1.80E+00	&1.58E+00	&\underline{1.56E+00}	&1.63E+00	&1.59E+00	&1.68E+00	&\underline{\textbf{1.49E+00}}        \\
    \cmidrule(l){2-11}
    & ETTh2   & 24
    &6.03E+00	&9.56E+00	&5.42E+00	&5.49E+00	&5.46E+00	&5.64E+00	&\underline{4.58E+00}	&\underline{\textbf{4.08E+00}}    \\
    \midrule

    \multirow{10}{*}{MAPE} 
    & \multirow{2}{*}{MG} & 17 
    &2.54E-01	&8.88E-03	&4.33E-03	&4.60E-03	&1.41E-02	&5.65E-03	&\underline{1.23E-03}  &\underline{\textbf{1.15E-03}} \\
    \cmidrule(l){3-11}
    &   & 84 
    &3.29E-01	&1.56E-02	&\underline{5.47E-03}	&8.75E-02	&1.86E-01	&1.06E-01	&9.79E-02	& \underline{\textbf{2.17E-03}}  \\
    \cmidrule(l){2-11}

    & \multirow{2}{*}{SML} & 48 
    &1.07E-01	&6.11E-02	&\underline{5.59E-02}	&6.84E-02	&9.37E-02	&7.03E-02	&8.26E-02	&\underline{\textbf{4.21E-02}}\\
    \cmidrule(l){3-11}
    &   & 96 
    &1.15E-01	&6.46E-02	&\underline{5.88E-02}	&9.52E-02	&9.89E-02	&9.56E-02	&1.16E-01	&\underline{\textbf{4.26E-02}}\\
    \cmidrule(l){2-11}

    & \multirow{2}{*}{PM} &30 
    &5.81E-01	&7.00E-01	&5.95E-01	&5.07E-01	&\underline{\textbf{4.73E-01}}	&5.09E-01	&5.18E-01	&\underline{5.01E-01}        \\
    \cmidrule(l){3-11}
    &   & 60 
    &6.40E-01	&7.34E-01	&5.79E-01	&5.54E-01	&\underline{5.37E-01}	&5.62E-01	&5.63E-01	&\underline{\textbf{5.20E-01}}    \\
    \cmidrule(l){2-11}

    & \multirow{2}{*}{ILI} & 4 
    &1.28E-01	&1.34E-01	&1.58E-01	&1.14E-01	&\underline{\textbf{1.12E-01}}	&1.14E-01	&\underline{\textbf{1.12E-01}}	&\underline{1.13E-01}        \\
    \cmidrule(l){3-11}
    &   & 12 
    &2.18E-01	&\underline{1.63E-01}	&2.01E-01	&2.21E-01	&2.17E-01	&2.24E-01	&2.03E-01	&\underline{\textbf{1.78E-01}}    \\
    \cmidrule(l){2-11}
    & ETTh1 & 24
    &1.43E-01	&1.55E-01	&\underline{1.39E-01}	&\underline{1.39E-01}	&1.48E-01	&1.43E-01	&1.40E-01	&\underline{\textbf{1.32E-01}}        \\
    \cmidrule(l){2-11}
    &ETTh2   & 24  
    &1.31E-01	&2.06E-01	&1.14E-01	&1.14E-01	&\underline{1.09E-01}	&1.15E-01	&9.43E-02	&\underline{\textbf{8.77E-02}}    \\
    \midrule

    \multirow{10}{*}{SMAPE} 
    & \multirow{2}{*}{MG} & 17 
    &1.19E-01	&4.43E-03	&2.16E-03	&2.30E-03	&7.02E-03	&2.83E-03	&\underline{6.17E-04}	&\underline{\textbf{5.74E-04}}        \\
    \cmidrule(l){3-11}
    &   & 84 
    &1.53E-01	&7.75E-03	&\underline{2.73E-03}	&4.04E-02	&8.85E-02	&4.84E-02	&4.83E-02	&\underline{\textbf{1.08E-03}}    \\
    \cmidrule(l){2-11}
    
    & \multirow{2}{*}{SML} & 48 
    &5.45E-02	&3.10E-02	&\underline{2.79E-02}	&3.40E-02	&4.57E-02	&3.50E-02	&3.99E-02	&\underline{\textbf{2.10E-02}}\\
    \cmidrule(l){3-11}
    &   & 96 
    &5.75E-02	&3.28E-02	&\underline{2.98E-02}	&4.72E-02	&4.78E-02	&4.75E-02	&5.47E-02	&\underline{\textbf{2.11E-02}}\\
    \cmidrule(l){2-11}

    & \multirow{2}{*}{PM} &30 
    &2.38E-01	&2.21E-01	&2.09E-01	&\underline{1.93E-01}	&\underline{\textbf{1.89E-01}}	&\underline{1.93E-01}	&1.94E-01	&\underline{\textbf{1.89E-01}}        \\
    \cmidrule(l){3-11}
    &   & 60 
    &2.52E-01	&2.29E-01	&\underline{2.00E-01}	&2.04E-01	&2.03E-01	&2.06E-01	&2.06E-01	&\underline{\textbf{1.93E-01}}    \\
    \cmidrule(l){2-11}

    & \multirow{2}{*}{ILI} & 4 
    &6.18E-02	&6.71E-02	&8.02E-02	&5.75E-02	&\underline{\textbf{5.68E-02}}	&5.75E-02	&\underline{5.71E-02}	&\underline{5.71E-02}        \\
    \cmidrule(l){3-11}
    &   & 12 
    &9.95E-02	&\underline{\textbf{8.52E-02}}	&9.69E-02	&1.02E-01	&1.00E-01	&1.04E-01	&9.66E-02	&\underline{8.70E-02}    \\
    \cmidrule(l){2-11}
    &  ETTh1  & 24 
    &6.82E-02	&7.73E-02	&6.72E-02	&\underline{6.27E-02}	&6.54E-02	&6.38E-02	&6.43E-02	&\underline{\textbf{5.97E-02}}        \\
    \cmidrule(l){2-11}
    & ETTh2 & 24
    &6.54E-02	&1.19E-01	&5.98E-02	&5.92E-02	&5.73E-02	&6.02E-02	&\underline{4.81E-02}	&\underline{\textbf{4.40E-02}}    \\

    \bottomrule
    \end{tabular}
    \label{tab:tables.mainR_all.tex}
\end{table*}
Since normalization is a standard requirement for time series modeling and prediction \citep{2014Multi}, we scaled each series by the min-max method and divided it into training, validation and testing sets with a ratio of 0.64, 0.16 and 0.20, respectively \citep{2020Error}. In this study, we used random search to find the best hyperparameters \citep{2012Random}, for the MSVR and MLP models constructed as in \citep{yangComprehensiveLearningParticle2021a}. For S2S models, since the optimization process is very time-consuming and this study focuses on the comparison among different training approaches, the general hyperparameters, for example, the unit dimensions of the RNNs, were determined according to the same rules of thumb to ensure the consistency of the initial untrained S2S network and improve the comparability of results. Only the specific hyperparameters for each training approach were optimized by random search, \emph{e.g.}, the schedule to decrease the probability \(p\) of choosing the true values as inputs in schedule sampling, the parameters of the distinct neural network in professor forcing, and the parameters of the policy network in PG-S2S. After determining the hyperparameters, we constructed and trained the models with the training set. Finally, we tested the trained models on the testing set and computed the accuracy metrics, \emph{i.e.} RMSE, MAPE, and SMAPE. The training and testing processes were executed ten times for each model independently to ensure the reliability of the results.

\section{Results and Discussion}
\subsection{Comparison on Prediction performance}
The prediction performances of all the benchmark models were examined in terms of three accuracy measures. The averaged results for each model on different series are reported in Table \ref{tab:tables.mainR_all.tex}. For each accuracy measure (RMSE, MAPE, and SMAPE), the entry with the smallest value of each row is set in boldface and underlined. The second-smallest value of each row is underlined only. 
The results in Table \ref{tab:tables.mainR_all.tex} lead to the following conclusions.
\begin{itemize}
\item{The proposed PG-S2S outperforms other competing models and training approaches in both univariate and multivariate tasks across different prediction horizons. The advantages of the proposed approach become more significant relative to other training approaches when any one of the auxiliary models performs well, such as H=84 over MG series and H=48 or 96 over SML series. It is conceivable that one reason for the superiority of the PG-S2S is its ability to effectively use predictive information from the auxiliary models by introducing reinforcement learning.}
\item {Despite the popularity of the teacher forcing in multi-step-ahead prediction literature, its performance does not consistently surpass that of the free running approach due to exposure bias, \emph{e.g.}, on MG series, SML series, and ETTh1 series, which is in agreement with \citep{gasparinDeepLearningTime2019,sangiorgioForecastingNoisyChaotic2021_FR}. This is another conceivable reason for the superiority of the proposed approach that mitigates exposure bias by feeding selected more accurate predictions from auxiliary models.}
\item {Moreover, in tasks where the free running approach produces predictions that are more accurate than those of the teacher forcing approach, this advantage diminishes as the forecasting horizon increases. 
This is because the cumulative error of the free running approach becomes much larger as the forecasting horizon increases, while teacher forcing, in which ground-truth values are fed into the decoder, acts as a certain correction and blocks the rapid amplification of errors. }
\end{itemize}

\subsection{Analysis of the Effectiveness of Reinforcement Learning}
To further assess the effectiveness of the proposed approach, we carried out a comparative analysis with two others without reinforcement learning.
\begin{itemize}
        \item \textbf{Teach\_MSVR}. During both training and testing, the decoder always uses predictions from the well-trained MSVR model as inputs and guide to decode the predictions for the next time steps.
        \item \textbf{Teach\_MLP}. Similarly, the decoder always uses predictions from the well-trained MLP model during both training and testing.
\end{itemize}

The results of these methods for all prediction tasks are reported in Table \ref*{tab:tables.agentR.tex}. For each accuracy measure, the entry with the smallest value in each column is highlighted with boldface and underlines. The PG-S2S keeps excelling other methods, providing preliminary evidence for the validity of reinforcement learning.
\begin{table*}[!t]
    \footnotesize
    \caption{Results of Comparative methods on the Five Datasets}
    \centering
    \footnotesize
    \setlength{\tabcolsep}{2pt}
    \begin{tabular}{ccccccccccccc}
    \toprule
    \multirow{2}{*}{Metric} & \multirow{2}{*}{Model} & & \multicolumn{2}{c}{MG}  & \multicolumn{2}{c}{SML} & \multicolumn{2}{c}{PM} & \multicolumn{2}{c}{ILI} &ETTh1 & ETTh2\\
    \cmidrule(l){4-5} \cmidrule(l){6-7} \cmidrule(l){8-9} \cmidrule(l){10-11} \cmidrule(l){12-12} \cmidrule(l){13-13}
    &  & &17 &84 &48 &96 &30 &60 &4 &12 &24 &24 \\
    \midrule

    \multirow{6}{*}{RMSE} & \multicolumn{2}{l}{\multirow{2}{*}{Teach\_MSVR}} 
    &6.58E-03	&1.13E-02	&1.11E+00	&1.20E+00	&\textbf{\underline{3.59E+01}}	&3.65E+01	&6.75E-01	&1.09E+00	&1.66E+00	&6.95E+00  \\
    & \multicolumn{2}{l}{}                  
    &\tiny{(\(\pm\)1.10E-04)}	&\tiny{(\(\pm\)5.03E-04)}	&\tiny{(\(\pm\)7.18E-02)}	&\tiny{(\(\pm\)8.84E-02)}	&\tiny{(\(\pm\)4.03E-01)}	&\tiny{(\(\pm\)4.45E-01)}	&\tiny{(\(\pm\)1.72E-02)}	&\tiny{(\(\pm\)1.50E-01)}	&\tiny{(\(\pm\)2.32E-01)}	&\tiny{(\(\pm\)4.04E-01)}\\  
    \cmidrule(l){2-13}

    & \multicolumn{2}{l}{\multirow{2}{*}{Teach\_MLP}} 
    &4.64E-03	&6.72E-03	&1.08E+00	&1.15E+00	&3.66E+01	&3.64E+01	&6.71E-01	&1.03E+00	&1.60E+00	&4.28E+00  \\
    & \multicolumn{2}{l}{}                  
    &\tiny{(\(\pm\)2.01E-04)}	&\tiny{(\(\pm\)1.01E-03)}	&\tiny{(\(\pm\)7.18E-02)}	&\tiny{(\(\pm\)8.84E-02)}	&\tiny{(\(\pm\)8.86E-01)}	&\tiny{(\(\pm\)3.09E-01)}	&\tiny{(\(\pm\)1.65E-02)}	&\tiny{(\(\pm\)4.00E-02)}	&\tiny{(\(\pm\)1.11E-01)}	&\tiny{(\(\pm\)1.28E-01)}\\  
    \cmidrule(l){2-13}

    & \multicolumn{2}{l}{\multirow{2}{*}{PG-S2S}} 
    &\textbf{\underline{1.25E-03}}	&\textbf{\underline{2.55E-03}}	&\textbf{\underline{1.04E+00}}	&\textbf{\underline{1.10E+00}}	&\textbf{\underline{3.59E+01}}	&\textbf{\underline{3.61E+01}}	&\textbf{\underline{6.68E-01}}	&\textbf{\underline{9.71E-01}}	&\textbf{\underline{1.49E+00}}	&\textbf{\underline{4.08E+00}}  \\
    & \multicolumn{2}{l}{}                  
    &\tiny{(\(\pm\)8.74E-05)}	&\tiny{(\(\pm\)9.50E-04)}	&\tiny{(\(\pm\)4.84E-02)}	&\tiny{(\(\pm\)2.85E-02)}	&\tiny{(\(\pm\)3.99E-01)}	&\tiny{(\(\pm\)3.95E-01)}	&\tiny{(\(\pm\)8.44E-03)}	&\tiny{(\(\pm\)1.83E-02)}	&\tiny{(\(\pm\)2.19E-02)}	&\tiny{(\(\pm\)6.75E-02)}\\  
    \midrule

    \multirow{6}{*}{MAPE} & \multicolumn{2}{l}{\multirow{2}{*}{Teach\_MSVR}} 
    &6.65E-03	&1.11E-02	&4.89E-02	&5.15E-02	&5.50E-01	&5.39E-01	&1.15E-01	&2.41E-01	&1.49E-01	&1.40E-01  \\
    & \multicolumn{2}{l}{}                  
    &\tiny{(\(\pm\)1.07E-04)}	&\tiny{(\(\pm\)4.53E-04)}	&\tiny{(\(\pm\)4.14E-03)}	&\tiny{(\(\pm\)5.05E-03)}	&\tiny{(\(\pm\)1.25E-02)}	&\tiny{(\(\pm\)1.76E-02)}	&\tiny{(\(\pm\)2.91E-03)}	&\tiny{(\(\pm\)6.78E-02)}	&\tiny{(\(\pm\)2.53E-02)}	&\tiny{(\(\pm\)9.84E-03)}\\  
    \cmidrule(l){2-13}

    & \multicolumn{2}{l}{\multirow{2}{*}{Teach\_MLP}} 
    &4.22E-03	&5.89E-03	&\textbf{\underline{4.20E-02}}	&4.44E-02	&5.16E-01	&5.52E-01	&1.14E-01	&2.15E-01	&1.44E-01	&9.20E-02  \\
    & \multicolumn{2}{l}{}                  
    &\tiny{(\(\pm\)1.88E-04)}	&\tiny{(\(\pm\)6.37E-04)}	&\tiny{(\(\pm\)3.91E-03)}	&\tiny{(\(\pm\)3.94E-03)}	&\tiny{(\(\pm\)1.16E-02)}	&\tiny{(\(\pm\)9.63E-03)}	&\tiny{(\(\pm\)3.01E-03)}	&\tiny{(\(\pm\)4.40E-02)}	&\tiny{(\(\pm\)1.27E-02)}	&\tiny{(\(\pm\)2.86E-03)}\\  
    \cmidrule(l){2-13}

    & \multicolumn{2}{l}{\multirow{2}{*}{PG-S2S}} 
    &\textbf{\underline{1.15E-03}}	&\textbf{\underline{2.17E-03}}	&4.21E-02	&\textbf{\underline{4.26E-02}}	&\textbf{\underline{5.01E-01}}	&\textbf{\underline{5.20E-01}}	&\textbf{\underline{1.13E-01}}	&\textbf{\underline{1.78E-01}}	&\textbf{\underline{1.32E-01}}	&\textbf{\underline{8.77E-02}}  \\
    & \multicolumn{2}{l}{}                  
    &\tiny{(\(\pm\)8.31E-05)}	&\tiny{(\(\pm\)7.36E-04)}	&\tiny{(\(\pm\)4.20E-03)}	&\tiny{(\(\pm\)2.69E-03)}	&\tiny{(\(\pm\)2.56E-02)}	&\tiny{(\(\pm\)2.61E-02)}	&\tiny{(\(\pm\)1.47E-03)}	&\tiny{(\(\pm\)2.17E-02)}	&\tiny{(\(\pm\)2.47E-03)}	&\tiny{(\(\pm\)1.41E-03)}\\  
    \midrule

    \multirow{6}{*}{SMAPE} & \multicolumn{2}{l}{\multirow{2}{*}{Teach\_MSVR}} 
    &3.32E-03	&5.55E-03	&2.40E-02	&2.53E-02	&1.93E-01	&1.96E-01	&5.78E-02	&1.12E-01	&6.60E-02	&7.75E-02  \\
    & \multicolumn{2}{l}{}                  
    &\tiny{(\(\pm\)5.45E-05)}	&\tiny{(\(\pm\)2.23E-04)}	&\tiny{(\(\pm\)1.92E-03)}	&\tiny{(\(\pm\)2.34E-03)}	&\tiny{(\(\pm\)1.85E-03)}	&\tiny{(\(\pm\)2.89E-03)}	&\tiny{(\(\pm\)1.59E-03)}	&\tiny{(\(\pm\)2.98E-02)}	&\tiny{(\(\pm\)9.22E-03)}	&\tiny{(\(\pm\)6.23E-03)}\\  
    \cmidrule(l){2-13}

    & \multicolumn{2}{l}{\multirow{2}{*}{Teach\_MLP}} 
    &2.11E-03	&2.95E-03	&2.11E-02	&2.20E-02	&1.93E-01	&1.94E-01	&5.75E-02	&1.00E-01	&6.46E-02	&4.65E-02  \\
    & \multicolumn{2}{l}{}                  
    &\tiny{(\(\pm\)9.45E-05)}	&\tiny{(\(\pm\)3.19E-04)}	&\tiny{(\(\pm\)1.86E-03)}	&\tiny{(\(\pm\)1.98E-03)}	&\tiny{(\(\pm\)2.20E-03)}	&\tiny{(\(\pm\)1.83E-03)}	&\tiny{(\(\pm\)1.63E-03)}	&\tiny{(\(\pm\)1.41E-02)}	&\tiny{(\(\pm\)5.66E-03)}	&\tiny{(\(\pm\)1.87E-03)}\\  
    \cmidrule(l){2-13}

    & \multicolumn{2}{l}{\multirow{2}{*}{PG-S2S}} 
    &\textbf{\underline{5.74E-04}}	&\textbf{\underline{1.08E-03}}	&\textbf{\underline{2.10E-02}}	&\textbf{\underline{2.11E-02}}	&\textbf{\underline{1.89E-01}}	&\textbf{\underline{1.93E-01}}	&\textbf{\underline{5.71E-02}}	&\textbf{\underline{8.70E-02}}	&\textbf{\underline{5.97E-02}}	&\textbf{\underline{4.40E-02}}    \\
    & \multicolumn{2}{l}{}                  
    &\tiny{(\(\pm\)4.16E-05)}	&\tiny{(\(\pm\)3.68E-04)}	&\tiny{(\(\pm\)1.89E-03)}	&\tiny{(\(\pm\)1.28E-03)}	&\tiny{(\(\pm\)2.24E-03)}	&\tiny{(\(\pm\)3.63E-03)}	&\tiny{(\(\pm\)7.67E-04)}	&\tiny{(\(\pm\)7.27E-03)}	&\tiny{(\(\pm\)1.49E-03)}	&\tiny{(\(\pm\)8.39E-04)}\\  
 
    \bottomrule
    \end{tabular}
    \label{tab:tables.agentR.tex}
\end{table*}

Next, we also explored how the dynamic, adaptive features unique to reinforcement learning take effect in input selection for the decoder with regard to the training, validation, and testing processes respectively.

\begin{figure}[!t]
	\centering
	\subfloat[ILI H4]{
		\label{fig:T_ILI.H4}
		\begin{minipage}[c]{0.40\textwidth}
		\centering
		\includegraphics[width=\textwidth]{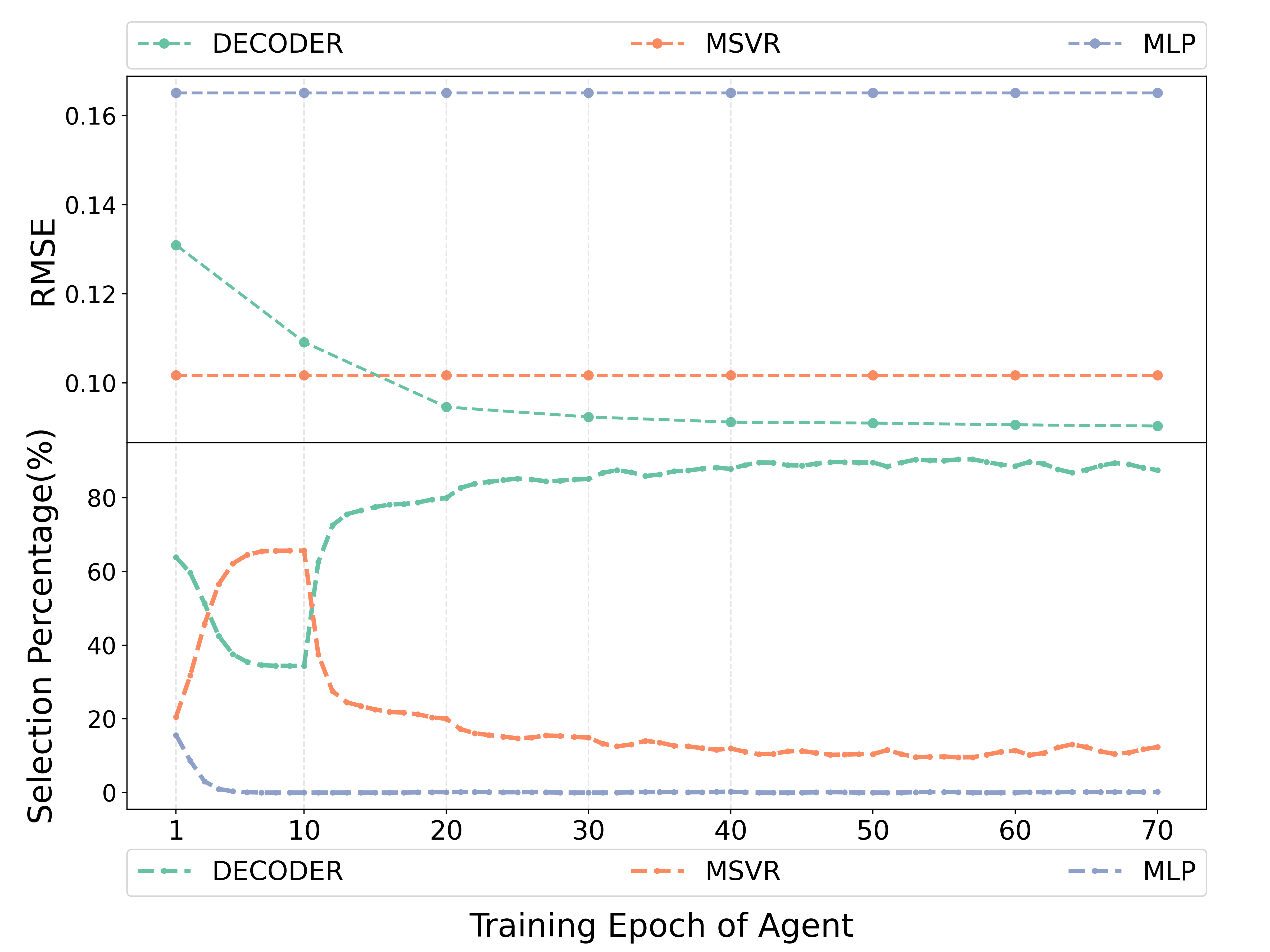}
		\end{minipage}
	}
	\quad
	\subfloat[ETTh1 H24]{
		\label{fig:T_ETTh1.H24}
		\begin{minipage}[c]{0.40\textwidth}
		\centering
		\includegraphics[width=\textwidth]{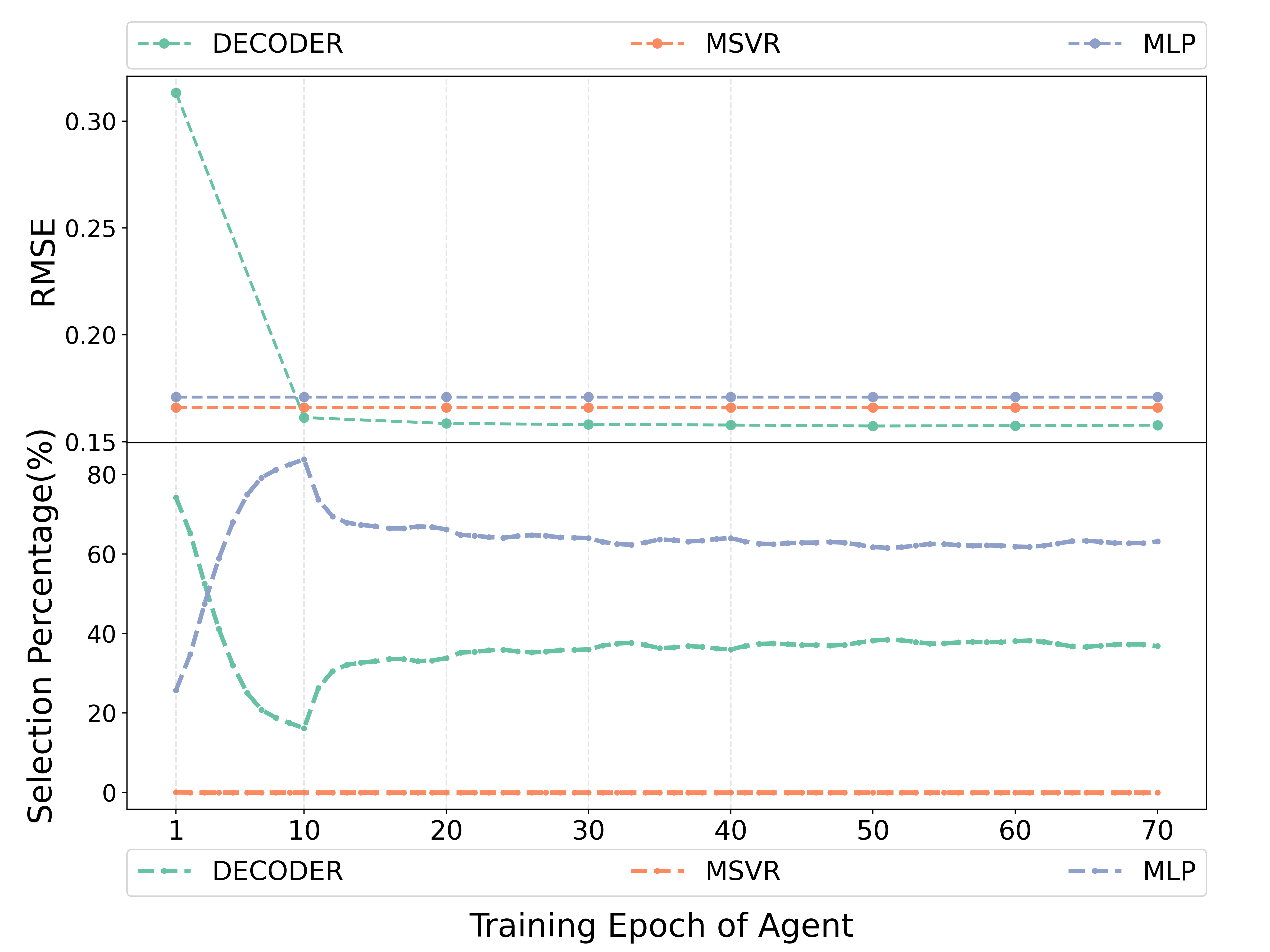}
		\end{minipage}
	}
	\caption{Training process of agent on different prediction tasks.}
	\label{fig:agentTrain}
\end{figure}

Fig. \ref{fig:agentTrain} illustrates the changes in the actions taken by the agent during the training on tasks of H=4 over the ILI series and H=24 over the ETTh1 series. The parameters of PG-S2S are updated asynchronously as described in Algorithm \ref{alg:alg1}, and the agent is trained for only 10 epochs per round. After each training round, the agent is fixed and enters a waiting state until the S2S network has completed a training round before proceeding to the next one. The upper half of each figure displays the RMSE of models in the pool at the beginning of each training round, where the auxiliary models have been well trained in advance and remain unchanged throughout, while that of the decoder gradually decreases with the S2S network training. The lower half shows the changes in the actions taken by the agent at the first time step, \emph{i.e.}, \(a_1\), during each training round. The selection percentage of each model is calculated as the proportion of all samples in the training set for which the agent selects this model at the first time step.

Let's have a close look with lens to understand how the proposed PG-S2S works at Fig. \ref{fig:agentTrain} as an illustration. In Fig. \ref{fig:agentTrain}\subref{fig:T_ILI.H4}, the decoder initially exhibits poor performance, with MSVR being the superior model. Consequently, during the first training round, the agent increasingly selects MSVR for a larger proportion of samples at \(a_1\), while concurrently decreasing its selection of the decoder. At the onset of the second training round, the prediction accuracy of the decoder, having undergone one training round with the predictions of the selected models as inputs, becomes commensurate with that of MSVR. Accordingly, the agent diminishes its predilection for MSVR and augments its preference for the decoder. By the third training round, the decoder surpasses MSVR in prediction accuracy, resulting in a further rise in the decoder's preference, which eventually stabilizes at approximately 85\% after several training rounds. A similar pattern can be observed in Fig. \ref{fig:agentTrain}\subref{fig:T_ETTh1.H24}. It is reasonable to conclude that the agent is capable of dynamically and adaptively updating its policy to select the most suitable models from the candidates to provide inputs for the decoder, thereby enhancing its prediction accuracy, even though the optimal model in the pool is not set in stone during the training process. Moreover, this also validates the effectiveness of the asynchronous updating strategy.

\begin{figure}[!t]
	\centering
    \subfloat{
            \label{fig:TV_ETTh2.H24}
            \begin{minipage}[c]{0.40\textwidth}
            \centering
            \includegraphics[width=\textwidth]{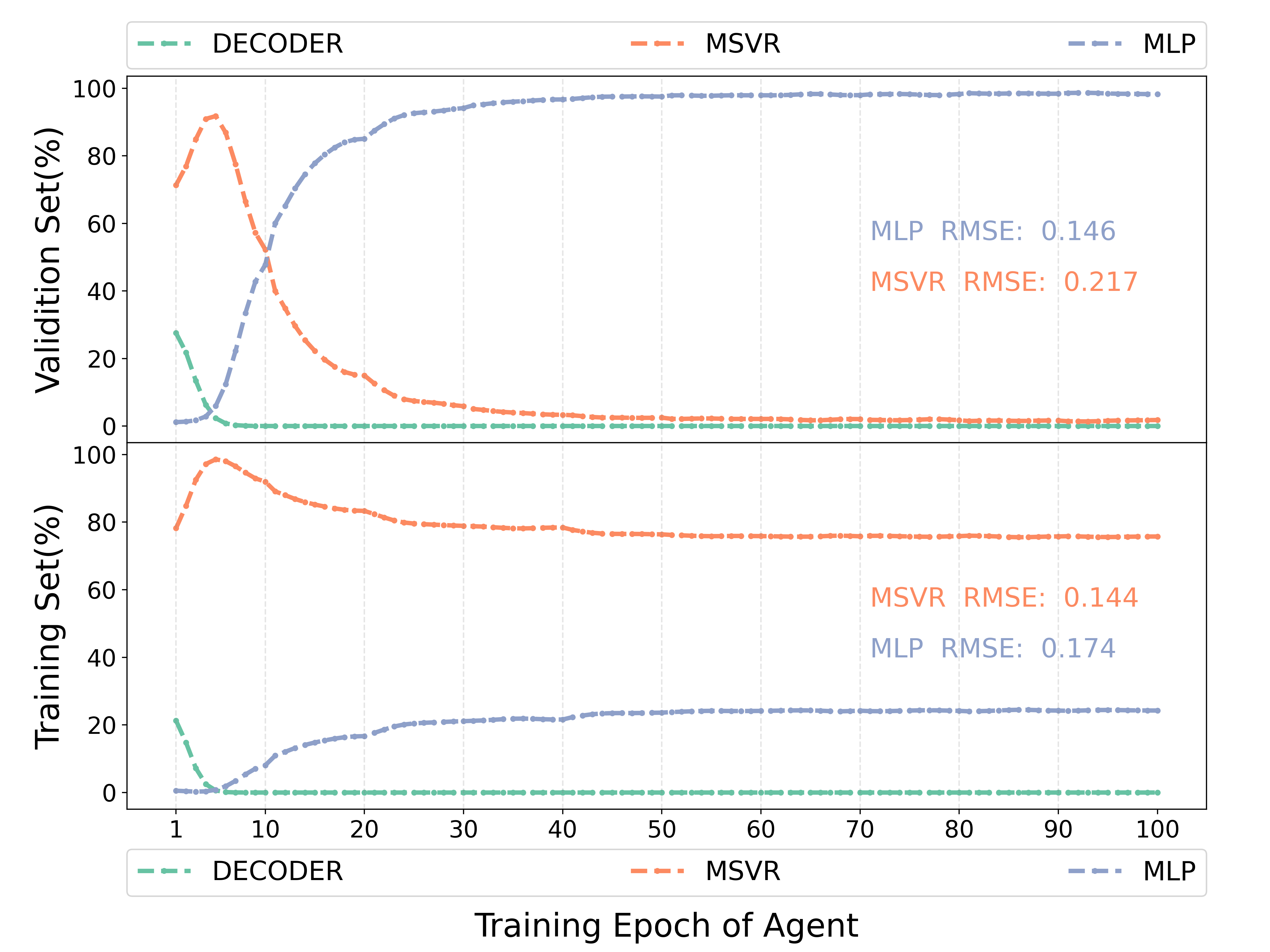}
            \end{minipage}
        }
	\caption{ETTh2-H24: The percentage of models in the training and validation datasets that are selected by the agent.}
	\label{fig:agentTV}
\end{figure}

Fig. \ref{fig:agentTV} compares the differences in model preference of the agent between the training and the validation sets over multiple training rounds. The samples in the validation set are not involved in updating the agent's policy. In the training set, the predictions of well-trained MLP (\text{RMSE}=0.174) are inferior to those of well-trained MSVR (\text{RMSE}=0.144). Therefore, throughout the training process, the agent selects MSVR for a larger percentage of samples. However, it is noteworthy that, in contrast to its behaviors on the training set, on the validation set, the percentage of samples in which the agent selects MSVR (\text{RMSE}=0.217) gradually diminishes and is significantly lower than that of MLP (\text{RMSE}=0.146) at the end of the training process. This suggests that the agent doesn't merely learn to execute a fixed set of actions, but rather learns to discern the predictive capabilities of each model for different samples and adaptively takes actions accordingly. The knowledge acquired by the agent from the training set enables it to select the superior model to provide inputs, even in the absence of knowledge regarding actual prediction performances on new samples, thereby accounting for why the PG-S2S performs better than Teach\_MSVR, Teach\_MLP, and Free running.

\vspace{-5mm}
\subsection{Analysis of the Generalization of PG-S2S on Various RNN Structures}
The proposed approach is generic and can be applied to S2S models configured with various types of RNN units in addition to LSTM. With respect to testing the applicability of the proposed approach for S2S structures, a comparative experiment based on two other common RNN units (\emph{i.e.}, ERNN and GRU), was performed on the ILI dataset (H=12). Table \ref*{tab:transfer_R} shows the results, with the entry with the smallest value in each row highlighted in bold and underlined. The S2S model based on the PG-S2S achieves the best performance among various training approaches regardless of the type of units it is made up of, demonstrating its generality on different RNN structures.
\begin{table}[!t]
    \footnotesize
    \caption{Results of S2S Models Configured by Different Cells}
    \centering
    \footnotesize
    \setlength{\tabcolsep}{3.5pt}
    \begin{tabular}{ccccccc}
    \toprule
    \multirow{2}{*}{Cell} & \multirow{2}{*}{Metric} & \multicolumn{5}{c}{S2S models with different training approaches} \\ 
    \cmidrule(l){3-7}
    &  &\makecell{FR}&\makecell{TF} &\makecell{SS} & \makecell{PF} & \makecell{PG} \\
    \midrule

    \multirow{6}{*}{ERNN} & \multirow{2}{*}{RMSE} 
    &1.02E+00	&1.02E+00	&1.03E+00	&1.03E+00	&\underline{\textbf{9.87E-01}}\\
    &
    &\tiny{(\(\pm\)3.45E-02)}	&\tiny{(\(\pm\)4.80E-02)}	&\tiny{(\(\pm\)6.40E-02)}	&\tiny{(\(\pm\)3.96E-02)}	&\tiny{(\(\pm\)2.42E-02)}\\
    \cmidrule(l){2-7}
    & \multirow{2}{*}{MAPE} 
    &2.21E-01	&2.17E-01	&2.13E-01	&2.03E-01	&\underline{\textbf{2.01E-01}}\\
    &
    &\tiny{(\(\pm\)2.83E-02)}	&\tiny{(\(\pm\)3.58E-02)}	&\tiny{(\(\pm\)4.41E-02)}	&\tiny{(\(\pm\)4.37E-02)}	&\tiny{(\(\pm\)2.49E-02)}\\
    \cmidrule(l){2-7}
    & \multirow{2}{*}{SMAPE} 
    &1.02E-01	&1.00E-01	&9.99E-02	&9.66E-02	&\underline{\textbf{9.49E-02}}\\
    &
    &\tiny{(\(\pm\)1.00E-02)}	&\tiny{(\(\pm\)1.23E-02)}	&\tiny{(\(\pm\)1.48E-02)}	&\tiny{(\(\pm\)1.41E-02)}	&\tiny{(\(\pm\)8.08E-03)}\\
    \midrule 

    \multirow{6}{*}{GRU} & \multirow{2}{*}{RMSE} 
    &1.08E+00	&1.02E+00	&1.02E+00	&1.02E+00	&\underline{\textbf{9.73E-01}}\\
    &
    &\tiny{(\(\pm\)1.01E-01)}	&\tiny{(\(\pm\)4.61E-02)}	&\tiny{(\(\pm\)5.58E-02)}	&\tiny{(\(\pm\)6.84E-02)}	&\tiny{(\(\pm\)1.70E-02)}\\
    \cmidrule(l){2-7}
    & \multirow{2}{*}{MAPE} 
    &2.20E-01	&2.20E-01	&2.17E-01	&1.85E-01	&\underline{\textbf{1.74E-01}}\\
    &   
    &\tiny{(\(\pm\)4.87E-02)}	&\tiny{(\(\pm\)3.63E-02)}	&\tiny{(\(\pm\)3.84E-02)}	&\tiny{(\(\pm\)5.18E-02)}	&\tiny{(\(\pm\)1.15E-02)}    \\
    \cmidrule(l){2-7}
    & \multirow{2}{*}{SMAPE} 
    &1.04E-01	&1.01E-01	&9.97E-02	&9.23E-02	&\underline{\textbf{8.63E-02}} \\
    &
    &\tiny{(\(\pm\)1.76E-02)}	&\tiny{(\(\pm\)1.17E-02)}	&\tiny{(\(\pm\)1.27E-02)}	&\tiny{(\(\pm\)1.69E-02)}	&\tiny{(\(\pm\)4.02E-03)}\\
    
    \bottomrule
    \end{tabular}
    \label{tab:transfer_R}
\end{table}
\section{Conclusion}
The sequence-to-sequence (S2S) models based on recurrent neural networks (RNNs) have been applied to multi-step-ahead time series prediction and drawn more attention, recently. Free running and teacher forcing are two popular training approaches but have rapid error amplification and the exposure bias problem, respectively. In this study, we propose the PG-S2S, a novel training approach and model, which is innovative in constructing model pool and introducing reinforcement learning algorithms for decoder input selection to improve prediction accuracy and robustness. Experimental results and comparisons demonstrate the superiority of the proposed approach. Further experimental analysis evidences that the adaptive and dynamic nature of reinforcement learning plays a crucial role in the decoding process. In fact, with the exception of S2S, the other models for multi-step-ahead prediction based on autoregressive mechanism all have the same training problem, which will be investigated in the future.

\bibliography{myref}

\end{document}